\documentclass[runningheads]{llncs}
\usepackage[T1]{fontenc}
\usepackage{amsmath}
\usepackage{amssymb}
\usepackage{graphicx}
\usepackage{makecell}
\usepackage{enumitem}
\usepackage[colorlinks=true, allcolors=blue]{hyperref}
\begin{document}
\title{Augmentation Strategy for DINO in Glaucoma Classification}
%
%
\author{Ashish Kumar Meena$^\star$ \and Chandra Sekhar Seelamantula}
\authorrunning{A.K. Meena and C.S. Seelamantula}
\institute{Department of Electrical Engineering, \\
Indian Institute of Science, Bengaluru, India\\
\email{ashishkumarmeena256@gmail.com$^\star$, css@iisc.ac.in}}
\maketitle              
\begin{abstract}
Self-supervised learning (SSL) offers a promising approach to medical image representation learning, yet augmentation strategies of SSL frameworks validated on natural images do not reliably transfer to specialized medical imaging domains. We analyze DINO, a knowledge-distillation based SSL framework, and empirically demonstrate that its default augmentation strategy produces unreliable representations for retinal fundus images, degrading downstream glaucoma classification performance. Guided by the principle that effective cropped views should share minimal redundant information while retaining clinically relevant signal, we redesign DINO's augmentation strategy. Our approach biases the learned representations toward diagnostically critical features rather than low-level image statistics. We evaluate our method for glaucoma classification by training a linear classifier on the frozen representations, achieving strong generalization across seven benchmark datasets. URL for code and weights: \url{https://github.com/ashish-256/task_dino}

\end{abstract}
\section{Introduction}
Fundus images vary significantly across acquisition instruments and patient demographics, causing supervised deep learning models to capture dataset-specific artifacts rather than generalizable anatomical structure, leaving them vulnerable to domain shift~\cite{supervised2}. SSL addresses this problem by learning transferable representations directly from unlabeled data. Among SSL methods, DINO~\cite{dino} has shown that self-supervised Vision Transformers develop semantically meaningful features with strong linear probing performance, motivating its adaptation to fundus imaging for generalizable representation learning.

However, recent studies indicate that augmentation strategies validated on natural images for SSL frameworks do not reliably generalize to specific medical imaging domains~\cite{selfsupervised2,selfsupervised3}. Moreover, they reveal that no single augmentation strategy is universally optimal, as performance instead depends on the alignment between the pretext task, the imaging modality, and the target task~\cite{selfsupervised3}. We examine this misalignment in the fundus imaging domain for glaucoma classification, and find that DINO's multi-crop objective, which enforces consistency across all crops regardless of anatomical relevance, distorts task-relevant structure by suppressing features central to glaucoma classification (Section~\ref{analysis}).

To address this shortcoming, we redesign DINO's augmentation strategy through an asymmetric teacher-student crop assignment that aligns the pretext task with glaucoma relevant anatomy. 

Our contributions are as follows:

\begin{itemize}[itemsep=4pt, topsep=0pt, partopsep=0pt, parsep=0pt]
\item We identify a mismatch between the semantic consistency assumptions underlying DINO's multi-crop objective and the anatomical structure of retinal fundus images, and design a \textit{task-aware augmentation strategy} to address it.
\item We show that assigning \textit{low-redundancy, high task-relevance crops} to the teacher and \textit{multi-scale task-retaining crops} to the student consistently outperforms alternative teacher-student crop assignment strategies.
\item We evaluate our method on \textit{color fundus photography} for \textit{glaucoma classification}, showing that the learned representations capture anatomically meaningful structure and generalize strongly across diverse benchmarks.
\end{itemize}

\section{Related Works}

\subsubsection{Self-Supervised Learning on Medical Images}
Several works adopt SSL for medical image representation learning. MoCo-CXR \cite{mocock} applies contrastive SSL to chest X-rays with augmentation tailored to radiological structure. RETFound \cite{retfound} trains a large-scale retinal foundation model on fundus and OCT images using self-supervised pretraining strategies, while UNI \cite{uni} scales self-supervised pretraining across large histopathology datasets. Most medical SSL works directly transfer augmentation strategies from natural images without adapting them to domain-specific structural properties \cite{histopathology1,selfsupervised2,selfsupervised3}. We address this by redesigning DINO's augmentation strategy 
with domain-aware crops tailored to retinal fundus images.

\subsubsection{Augmentation Strategies for SSL}
The InfoMin principle \cite{tian2020makes} formalizes that optimal contrastive views balance task-relevant information against nuisance variability. In the medical imaging domain, FocusContrast \cite{gaze_paper_name} uses radiologist gaze to construct better contrastive views for knee X-rays, guiding crops toward diagnostically salient regions. In knowledge-distillation based SSL, DINO-LG \cite{dinolg} guides local crops toward annotated calcification regions for coronary calcium scoring, improving on unguided DINO. Similarly, DINO-EYE \cite{dinoeye} adapts DINO for retinal fundus images, demonstrating its utility in distinguishing optic disc phenotypes in primary open-angle glaucoma. However, neither DINO-LG nor DINO-EYE examine the asymmetric role of teacher and student views, specifically how cropped views should differ between them to produce task-aware representations. We address this by systematically evaluating teacher-student crop assignment strategies for DINO in the fundus imaging domain.

\begin{figure}
  \centering
  \includegraphics[width=1.0\columnwidth, trim=1cm 4cm 4cm 2.5cm, clip]{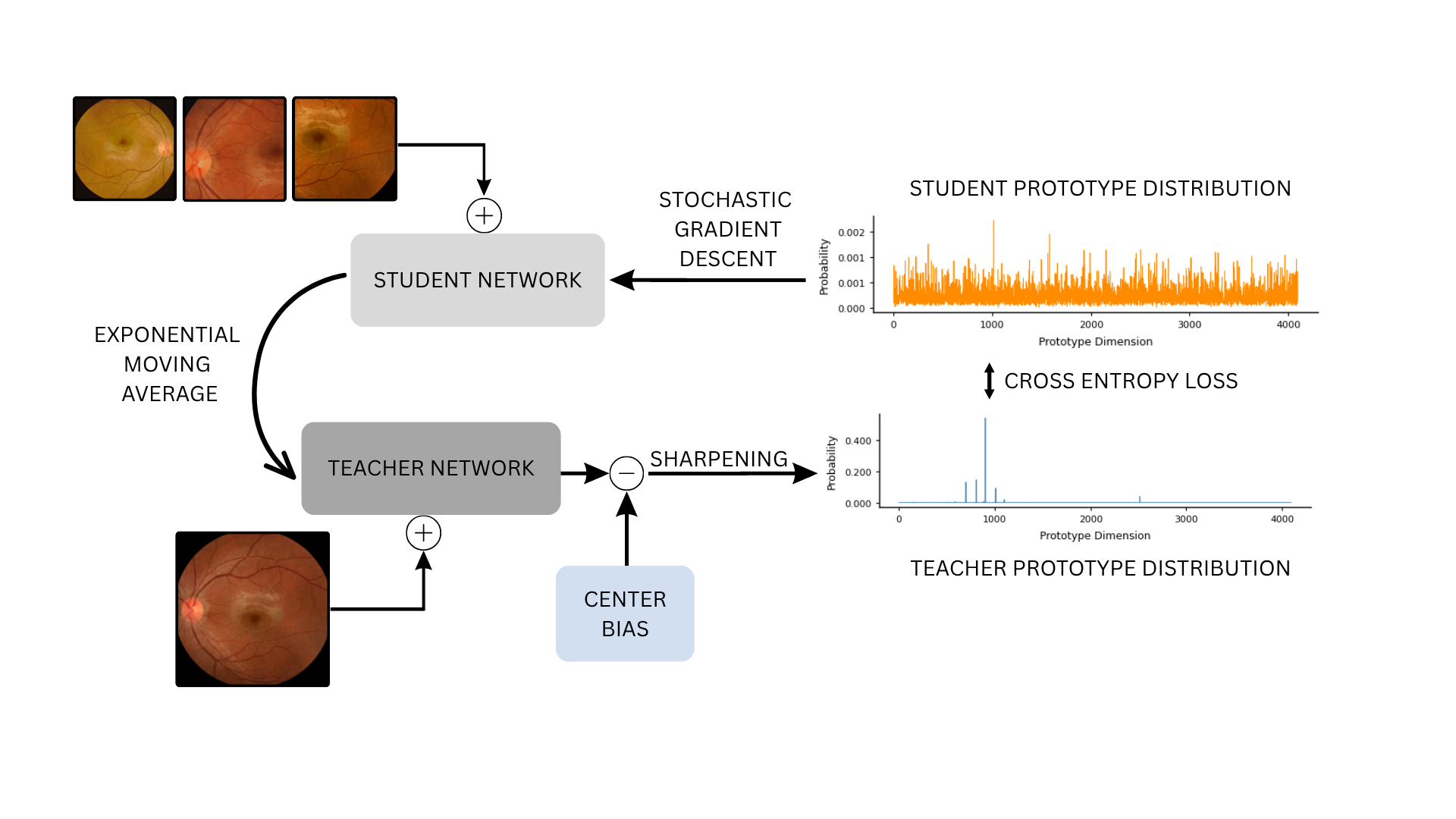}
  \caption{\textbf{Flow of information in DINO method:} Student learns augmentation invariant features by matching the teacher's target distribution across views.}
  \label{dino_explain}
\end{figure}

\section{Background: Analysis of DINO Architecture}
\label{analysis}
Vanilla DINO constructs representations through a teacher-student framework, where both 
networks share identical architectures: a ViT backbone followed by a projection head 
producing $K$-dimensional logits. During training, two global crops ($>50\%$ of the image) 
and several local crops ($<50\%$) are generated per image; the student processes all views 
while the teacher sees only global crops. Both outputs are converted to probability 
distributions via temperature-scaled softmax, and the student is trained to match the 
teacher's distributions using cross-entropy loss. The teacher receives no gradient updates; 
instead, its parameters are updated as an exponential moving average of the student weights, forming a temporally stable target. Representation collapse is prevented through centering, which subtracts a running mean from teacher outputs, and sharpening, which uses a lower temperature ($\tau_t < \tau_s$) to produce peaked distributions. 

DINO training can be interpreted as a positive feedback loop, as illustrated in Figure~\ref{dino_explain}. As the student improves, the EMA-updated teacher produces sharper, more confident targets, which in turn push the student toward progressively higher-order invariances, driving semantically similar images into clustered regions of the prototype space. This mechanism relies on the assumption that all crops of an image share a consistent semantic identity \cite{dino_survey}. In natural images, local crops typically retain sufficient semantic context for the student to infer the global identity of the scene, satisfying this assumption \cite{selfsupervised2}. Fundus image crops, however, are often clinically uninformative in isolation and may contain entirely different anatomical regions with no shared semantic correspondence, directly violating DINO's core assumption and undermining the reliability of the learned representations.

\begin{figure}
  \centering
  \includegraphics[width=1.0\linewidth, trim=5.4cm 1.7cm 5.4cm 1.7cm, clip]{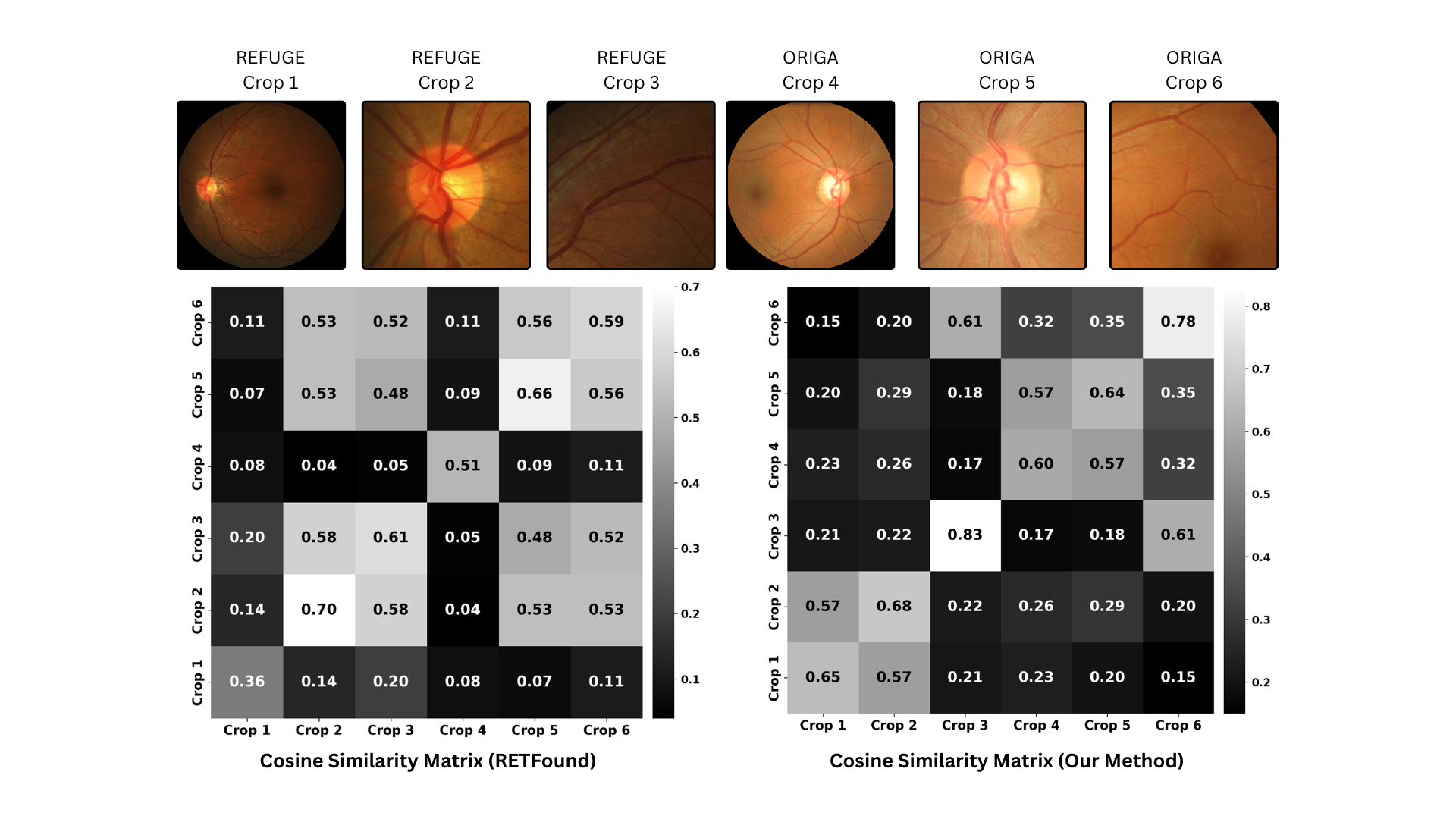}
  \caption{Enforcing embedding consistency across anatomically disjoint augmented crops drives the model to rely on shared low-level textural features, suppressing the learning of meaningful structural representations in fundus images.}
  \label{fig:cosine_similarity}
\end{figure}

\textbf{Empirical Validation:} To empirically validate the violation of DINO's semantic-consistency assumption in fundus imaging for glaucoma classification, we extract three crop types per image: full image, optic-disc crop, and a no-disc peripheral crop, each 224×224×3 (Figure~\ref{fig:cosine_similarity}). From each of REFUGE~\cite{refuge} and ORIGA~\cite{origa}, we sample 100 images and extract all three crop types from every image. For each crop-type pair, we compute the cosine similarity between all 100 CLS-token embeddings of one type and all 100 of the other, giving 10,000 pairwise comparisons per dataset, which we average into a single similarity score. Diagnostic signal for glaucoma is concentrated in the optic disc. A task-aware representation should therefore rate the disc crop as more similar to the full image (anatomically connected) than to the no-disc crop (anatomically disjoint). RETFound-DINOv2 ~\cite{retfound} shows the opposite in both datasets: disc to no-disc similarity (0.58/0.56) exceeds full to disc similarity (0.14/0.09), because DINO's multi-crop objective enforces consistency across all crops regardless of shared task-relevant content, an assumption that fails when crops fall outside the diagnostic region. Our task-aware augmentation strategy (Section ~\ref{taskaware}) recovers the correct ordering (full-to-disc: 0.57/0.57 vs. disc-to-no-disc: 0.22/0.35). 

\section{Methodology: Task-Aware Augmentation Strategy}
\label{taskaware}
Let $x$ denote an input fundus image and $y$ its task label (glaucoma status). In DINO, a teacher network $f_\xi$ and a student network $f_\theta$ each receive different augmented views of $x$: the teacher receives two views $v_T^{(j)}$, $j=1,2$, while the student receives $K$ views $v_S^{(k)}$, $k=1,\dots,K$. The student is trained to match the teacher's output distribution across its views, while $\xi$ is updated only via an exponential moving average (EMA) of $\theta$ and receives no gradient.

The InfoMin principle establishes that downstream task performance peaks at an intermediate level of mutual information (MI) between two such views, with the optimal level being task-dependent. Formally, given minimal sufficient encoders and label $y$, the optimal view pair $(v_1^*, v_2^*)$ satisfies
\begin{equation}
(v_1^*, v_2^*) = \arg\min_{v_1, v_2} I(v_1; v_2) \quad \text{subject to} \quad I(v_1; y) = I(v_2; y) = I(x; y),
\end{equation}
under which the representation learned by contrastive training on $(v_1^*, v_2^*)$ is provably optimal for the task~\cite{tian2020makes}. This formalizes the notion of a task-aware representation: one that minimizes shared nuisance information between views while preserving all information relevant to the downstream task, thereby improving generalization and reducing sample complexity.

Motivated by the InfoMin principle, we frame the teacher-student view pair $(v_T^{(j)}, v_S^{(k)})$ as one that should minimize their shared information while both remain sufficient for $y$:
\begin{equation}
\label{eq:infomin}
(v_T^{(j)*}, v_S^{(k)*}) = \arg\min_{v_T^{(j)},\, v_S^{(k)}} I(v_T^{(j)}; v_S^{(k)}) \;\; \text{s.t.} \;\; I(v_T^{(j)};y) = I(v_S^{(k)};y) = I(x;y),
\end{equation}
for all $j=1,2$, $k=1,\dots,K$, ensuring only task-relevant structure is shared between views. However, in standard contrastive learning, exchanging the two views does not affect the learning objective, since both pass through the same encoder. DINO breaks this symmetry: since $f_\xi \neq f_\theta$ and only $f_\theta$ receives gradients, the teacher acts as a self-generated label supervisor rather than a symmetric peer, defining target distributions the student is trained to match. This asymmetry determines how each view should be constructed. Both view types follow the same objective (Eq.~\ref{eq:infomin}), differing only in their permitted nuisance level $\delta$:
\begin{equation}
v^*(\delta) = \arg\max_v \; I(v;y) \quad \text{subject to} \quad I(v;x) - I(v;y) \leq \delta, \quad \delta \geq 0.
\end{equation}
Teacher views operate at $\delta_T \approx 0$, so $I(v_T;x) \approx I(v_T;y)$, meaning nearly all information $v_T$ retains about $x$ is task-relevant: $v_T$ approximates the minimal sufficient statistic of $x$ for $y$, producing reliable, task-grounded targets. Student views operate at $\delta_S \gg \delta_T$, retaining the broader structure of $x$: since only $f_\theta$ receives gradients and the final model is evaluated on images distributed like $x$, the student must learn to extract task-relevant representations from full, nuisance-retaining views. Since exact constraint satisfaction is unattainable in practice, we treat these as design motivation, trading a small loss in task-relevant information for a larger reduction in nuisance variation. 

\begin{figure}
  \centering
  \includegraphics[width=1.0\linewidth, trim=1cm 7cm 1cm 9cm, clip]{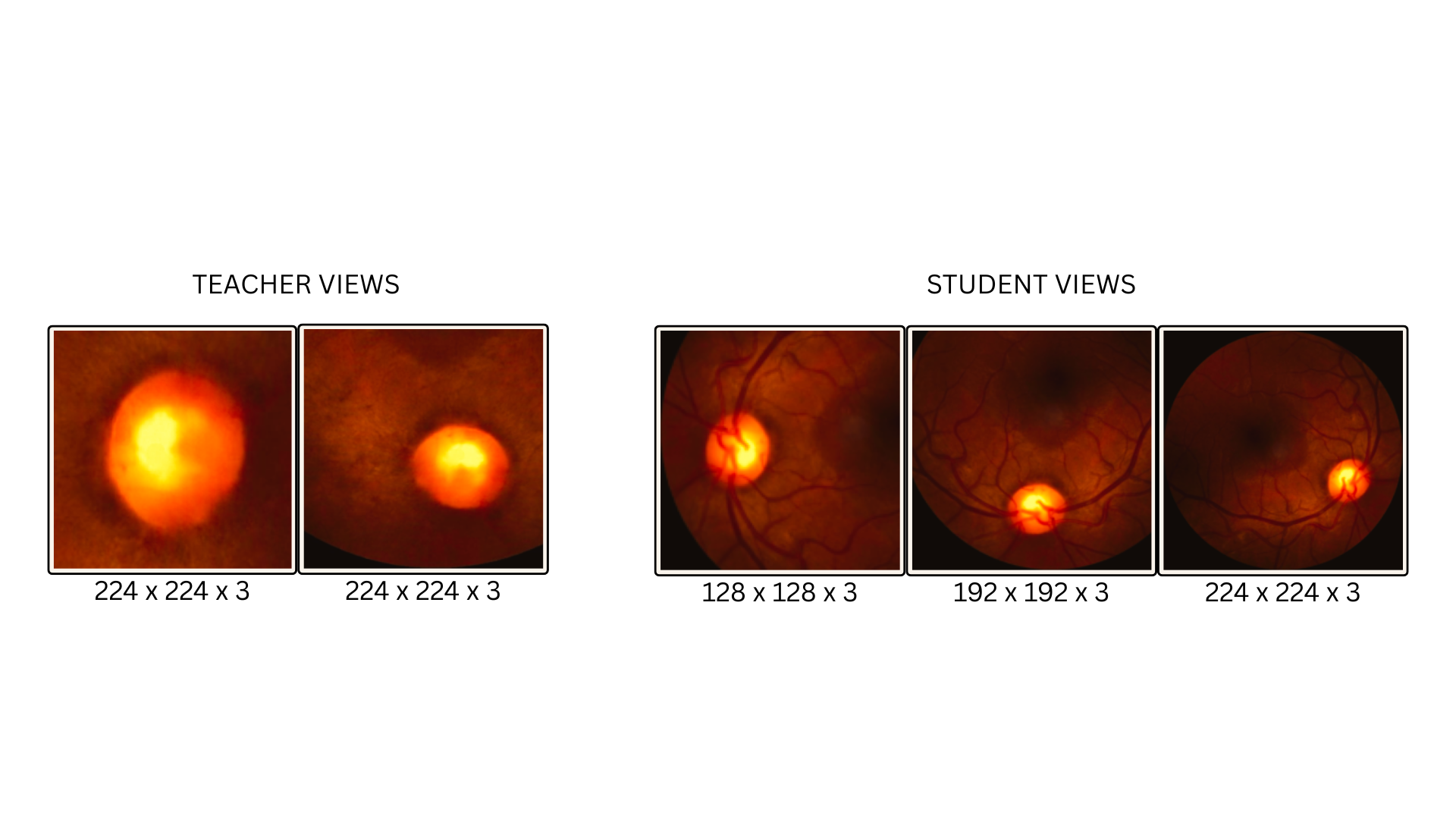}
  \caption{Proposed view assignment strategy for glaucoma classification}
  \label{fig:crop}
\end{figure}

\textbf{Augmentation Design for Glaucoma Classification:} Although vessel patterns carry some diagnostic signal for glaucoma, the primary diagnostic features lie in optic disc and cup morphology~\cite{vessel}. The teacher therefore receives two vessel-free, minimal-context crops of the optic disc. The student receives three multi-scale crops, each containing the optic disc (Figure~\ref{fig:crop}), encouraging scale-invariant representations while preserving task-relevant signal. All crops use color jitter, flips, and rotations, with Gaussian blur applied only to student crops for added view diversity. We validate our design choices against alternative crop-assignment strategies in Section~\ref{ablation}.

\section{Experiments and Results}
In this section, we describe our data preparation pipeline, training configuration, and evaluation protocol across seven glaucoma benchmarks, assessing representation quality via linear probing on frozen features. Ablation studies validate our augmentation design choices.

\textbf{Data Preparation:} For training DINO, we aggregate six publicly available datasets~\cite{hygd,airogs,fives,cataract,lag,origa}, filtering low-quality images via QuickQual~\cite{engelmann2023quickqual}. Optic disc localization uses a U-Net segmentation model, cross-validated against an independent model on mask overlap and circularity to discard poorly localized images. Vessel masks from a separate vessel segmentation model are used with LaMa~\cite{suvorov2022lama} to generate vessel-free images, with mask dilation and inpainting strength tuned conservatively to preserve the disc margin and blend inpainted regions with surrounding tissue. This pipeline yields a pretraining corpus of 81k fundus images with corresponding disc masks and vessel-free counterparts.

\textbf{Training Configuration:} We adopt ViT-Small (21M parameters) as the backbone, initialized from scratch, appropriate for our dataset scale, where larger backbones risk overfitting without sufficient data. The projection head produces 4096-dimensional prototypes, sufficient given the limited visual diversity of fundus images. Models are trained for 200 epochs with batch size 256, following the DINO recipe~\cite{dino} to prevent prototype collapse and ensure stable loss convergence.

\begin{table}[t]
\centering
\renewcommand{\arraystretch}{1.4}
\setlength{\tabcolsep}{4pt}
\caption{AUROC (95\% CIs) results under two linear-probe training regimes.}
\label{tab:results}
\resizebox{\textwidth}{!}{%
\begin{tabular}{|l|c|c|c|c|c|}
\hline
\textbf{Dataset} & \textbf{Vanilla DINO} & \textbf{RETFound} & \textbf{RetiZero} & \textbf{FLAIR} & \textbf{Ours} \\
\hline
\multicolumn{6}{|l|}{\textbf{(a) REFUGE-trained classifier ($\sim$1.2k images)}} \\
\hline
REFUGE &
  \shortstack{\rule{0pt}{2ex}0.802 \\ \scriptsize $[0.741$-$0.863]$\rule[-1ex]{0pt}{0pt}} &
  \shortstack{\rule{0pt}{2ex}0.875 \\ \scriptsize $[0.825$-$0.925]$\rule[-1ex]{0pt}{0pt}} &
  \shortstack{\rule{0pt}{2ex}\textbf{0.911} \\ \scriptsize $[0.876$-$0.947]$\rule[-1ex]{0pt}{0pt}} &
  \shortstack{\rule{0pt}{2ex}0.761 \\ \scriptsize $[0.701$-$0.821]$\rule[-1ex]{0pt}{0pt}} &
  \shortstack{\rule{0pt}{2ex}0.797 \\ \scriptsize $[0.739$-$0.856]$\rule[-1ex]{0pt}{0pt}} \\
\hline
ACRIMA &
  \shortstack{\rule{0pt}{2ex}0.824 \\ \scriptsize $[0.793$-$0.855]$\rule[-1ex]{0pt}{0pt}} &
  \shortstack{\rule{0pt}{2ex}0.824 \\ \scriptsize $[0.794$-$0.855]$\rule[-1ex]{0pt}{0pt}} &
  \shortstack{\rule{0pt}{2ex}0.778 \\ \scriptsize $[0.743$-$0.813]$\rule[-1ex]{0pt}{0pt}} &
  \shortstack{\rule{0pt}{2ex}0.717 \\ \scriptsize $[0.680$-$0.755]$\rule[-1ex]{0pt}{0pt}} &
  \shortstack{\rule{0pt}{2ex}\textbf{0.846} \\ \scriptsize $[0.818$-$0.874]$\rule[-1ex]{0pt}{0pt}} \\
\hline
LAG &
  \shortstack{\rule{0pt}{2ex}0.852 \\ \scriptsize $[0.841$-$0.864]$\rule[-1ex]{0pt}{0pt}} &
  \shortstack{\rule{0pt}{2ex}0.815 \\ \scriptsize $[0.801$-$0.828]$\rule[-1ex]{0pt}{0pt}} &
  \shortstack{\rule{0pt}{2ex}0.816 \\ \scriptsize $[0.803$-$0.829]$\rule[-1ex]{0pt}{0pt}} &
  \shortstack{\rule{0pt}{2ex}0.874 \\ \scriptsize $[0.862$-$0.885]$\rule[-1ex]{0pt}{0pt}} &
  \shortstack{\rule{0pt}{2ex}\textbf{0.890} \\ \scriptsize $[0.881$-$0.900]$\rule[-1ex]{0pt}{0pt}} \\
\hline
ORIGA &
  \shortstack{\rule{0pt}{2ex}0.692 \\ \scriptsize $[0.647$-$0.737]$\rule[-1ex]{0pt}{0pt}} &
  \shortstack{\rule{0pt}{2ex}0.694 \\ \scriptsize $[0.645$-$0.742]$\rule[-1ex]{0pt}{0pt}} &
  \shortstack{\rule{0pt}{2ex}0.776 \\ \scriptsize $[0.735$-$0.816]$\rule[-1ex]{0pt}{0pt}} &
  \shortstack{\rule{0pt}{2ex}\textbf{0.789} \\ \scriptsize $[0.748$-$0.829]$\rule[-1ex]{0pt}{0pt}} &
  \shortstack{\rule{0pt}{2ex}0.757 \\ \scriptsize $[0.716$-$0.798]$\rule[-1ex]{0pt}{0pt}} \\
\hline
FIVES &
  \shortstack{\rule{0pt}{2ex}0.835 \\ \scriptsize $[0.796$-$0.874]$\rule[-1ex]{0pt}{0pt}} &
  \shortstack{\rule{0pt}{2ex}0.804 \\ \scriptsize $[0.760$-$0.847]$\rule[-1ex]{0pt}{0pt}} &
  \shortstack{\rule{0pt}{2ex}0.847 \\ \scriptsize $[0.809$-$0.885]$\rule[-1ex]{0pt}{0pt}} &
  \shortstack{\rule{0pt}{2ex}0.852 \\ \scriptsize $[0.815$-$0.889]$\rule[-1ex]{0pt}{0pt}} &
  \shortstack{\rule{0pt}{2ex}\textbf{0.866} \\ \scriptsize $[0.831$-$0.901]$\rule[-1ex]{0pt}{0pt}} \\
\hline
PAPILA &
  \shortstack{\rule{0pt}{2ex}0.781 \\ \scriptsize $[0.730$-$0.833]$\rule[-1ex]{0pt}{0pt}} &
  \shortstack{\rule{0pt}{2ex}0.747 \\ \scriptsize $[0.685$-$0.809]$\rule[-1ex]{0pt}{0pt}} &
  \shortstack{\rule{0pt}{2ex}\textbf{0.869} \\ \scriptsize $[0.822$-$0.915]$\rule[-1ex]{0pt}{0pt}} &
  \shortstack{\rule{0pt}{2ex}0.703 \\ \scriptsize $[0.637$-$0.770]$\rule[-1ex]{0pt}{0pt}} &
  \shortstack{\rule{0pt}{2ex}0.786 \\ \scriptsize $[0.734$-$0.839]$\rule[-1ex]{0pt}{0pt}} \\
\hline
CHAKSU &
  \shortstack{\rule{0pt}{2ex}0.732 \\ \scriptsize $[0.695$-$0.770]$\rule[-1ex]{0pt}{0pt}} &
  \shortstack{\rule{0pt}{2ex}0.618 \\ \scriptsize $[0.572$-$0.663]$\rule[-1ex]{0pt}{0pt}} &
  \shortstack{\rule{0pt}{2ex}0.730 \\ \scriptsize $[0.693$-$0.768]$\rule[-1ex]{0pt}{0pt}} &
  \shortstack{\rule{0pt}{2ex}0.721 \\ \scriptsize $[0.680$-$0.762]$\rule[-1ex]{0pt}{0pt}} &
  \shortstack{\rule{0pt}{2ex}\textbf{0.784} \\ \scriptsize $[0.751$-$0.818]$\rule[-1ex]{0pt}{0pt}} \\
\hline
\textit{Mean $\pm$ Std} &
  $0.788 \pm 0.054$ & $0.768 \pm 0.082$ & $\mathbf{0.818 \pm 0.057}$ & $0.774 \pm 0.063$ & $\mathbf{0.818 \pm 0.046}$ \\
\hline
\multicolumn{6}{|l|}{\textbf{(b) AIROGS-trained classifier ($\sim$78k images)}} \\
\hline
REFUGE &
  \shortstack{\rule{0pt}{2ex}0.713 \\ \scriptsize $[0.645$-$0.780]$\rule[-1ex]{0pt}{0pt}} &
  \shortstack{\rule{0pt}{2ex}0.804 \\ \scriptsize $[0.740$-$0.868]$\rule[-1ex]{0pt}{0pt}} &
  \shortstack{\rule{0pt}{2ex}\textbf{0.848} \\ \scriptsize $[0.796$-$0.901]$\rule[-1ex]{0pt}{0pt}} &
  \shortstack{\rule{0pt}{2ex}0.789 \\ \scriptsize $[0.727$-$0.850]$\rule[-1ex]{0pt}{0pt}} &
  \shortstack{\rule{0pt}{2ex}0.792 \\ \scriptsize $[0.727$-$0.857]$\rule[-1ex]{0pt}{0pt}} \\
\hline
ACRIMA &
  \shortstack{\rule{0pt}{2ex}0.708 \\ \scriptsize $[0.670$-$0.747]$\rule[-1ex]{0pt}{0pt}} &
  \shortstack{\rule{0pt}{2ex}\textbf{0.817} \\ \scriptsize $[0.786$-$0.848]$\rule[-1ex]{0pt}{0pt}} &
  \shortstack{\rule{0pt}{2ex}0.698 \\ \scriptsize $[0.660$-$0.737]$\rule[-1ex]{0pt}{0pt}} &
  \shortstack{\rule{0pt}{2ex}0.798 \\ \scriptsize $[0.766$-$0.831]$\rule[-1ex]{0pt}{0pt}} &
  \shortstack{\rule{0pt}{2ex}0.792 \\ \scriptsize $[0.759$-$0.826]$\rule[-1ex]{0pt}{0pt}} \\
\hline
LAG &
  \shortstack{\rule{0pt}{2ex}0.893 \\ \scriptsize $[0.883$-$0.902]$\rule[-1ex]{0pt}{0pt}} &
  \shortstack{\rule{0pt}{2ex}0.857 \\ \scriptsize $[0.845$-$0.869]$\rule[-1ex]{0pt}{0pt}} &
  \shortstack{\rule{0pt}{2ex}0.913 \\ \scriptsize $[0.905$-$0.922]$\rule[-1ex]{0pt}{0pt}} &
  \shortstack{\rule{0pt}{2ex}0.899 \\ \scriptsize $[0.889$-$0.908]$\rule[-1ex]{0pt}{0pt}} &
  \shortstack{\rule{0pt}{2ex}\textbf{0.929} \\ \scriptsize $[0.921$-$0.936]$\rule[-1ex]{0pt}{0pt}} \\
\hline
ORIGA &
  \shortstack{\rule{0pt}{2ex}0.754 \\ \scriptsize $[0.714$-$0.794]$\rule[-1ex]{0pt}{0pt}} &
  \shortstack{\rule{0pt}{2ex}0.775 \\ \scriptsize $[0.733$-$0.817]$\rule[-1ex]{0pt}{0pt}} &
  \shortstack{\rule{0pt}{2ex}0.807 \\ \scriptsize $[0.770$-$0.845]$\rule[-1ex]{0pt}{0pt}} &
  \shortstack{\rule{0pt}{2ex}\textbf{0.820} \\ \scriptsize $[0.784$-$0.857]$\rule[-1ex]{0pt}{0pt}} &
  \shortstack{\rule{0pt}{2ex}0.787 \\ \scriptsize $[0.748$-$0.825]$\rule[-1ex]{0pt}{0pt}} \\
\hline
FIVES &
  \shortstack{\rule{0pt}{2ex}0.868 \\ \scriptsize $[0.832$-$0.904]$\rule[-1ex]{0pt}{0pt}} &
  \shortstack{\rule{0pt}{2ex}0.855 \\ \scriptsize $[0.819$-$0.892]$\rule[-1ex]{0pt}{0pt}} &
  \shortstack{\rule{0pt}{2ex}0.891 \\ \scriptsize $[0.858$-$0.923]$\rule[-1ex]{0pt}{0pt}} &
  \shortstack{\rule{0pt}{2ex}0.864 \\ \scriptsize $[0.827$-$0.890]$\rule[-1ex]{0pt}{0pt}} &
  \shortstack{\rule{0pt}{2ex}\textbf{0.923} \\ \scriptsize $[0.897$-$0.949]$\rule[-1ex]{0pt}{0pt}} \\
\hline
PAPILA &
  \shortstack{\rule{0pt}{2ex}0.808 \\ \scriptsize $[0.757$-$0.859]$\rule[-1ex]{0pt}{0pt}} &
  \shortstack{\rule{0pt}{2ex}0.835 \\ \scriptsize $[0.778$-$0.891]$\rule[-1ex]{0pt}{0pt}} &
  \shortstack{\rule{0pt}{2ex}\textbf{0.871} \\ \scriptsize $[0.828$-$0.914]$\rule[-1ex]{0pt}{0pt}} &
  \shortstack{\rule{0pt}{2ex}0.803 \\ \scriptsize $[0.746$-$0.859]$\rule[-1ex]{0pt}{0pt}} &
  \shortstack{\rule{0pt}{2ex}0.793 \\ \scriptsize $[0.740$-$0.846]$\rule[-1ex]{0pt}{0pt}} \\
\hline
CHAKSU &
  \shortstack{\rule{0pt}{2ex}0.717 \\ \scriptsize $[0.680$-$0.755]$\rule[-1ex]{0pt}{0pt}} &
  \shortstack{\rule{0pt}{2ex}0.667 \\ \scriptsize $[0.622$-$0.711]$\rule[-1ex]{0pt}{0pt}} &
  \shortstack{\rule{0pt}{2ex}0.796 \\ \scriptsize $[0.764$-$0.827]$\rule[-1ex]{0pt}{0pt}} &
  \shortstack{\rule{0pt}{2ex}0.766 \\ \scriptsize $[0.731$-$0.802]$\rule[-1ex]{0pt}{0pt}} &
  \shortstack{\rule{0pt}{2ex}\textbf{0.836} \\ \scriptsize $[0.808$-$0.864]$\rule[-1ex]{0pt}{0pt}} \\
\hline
\textit{Mean $\pm$ Std} &
  $0.780 \pm 0.071$ & $0.801 \pm 0.061$ & $0.832 \pm 0.067$ & $0.820 \pm 0.043$ & $\mathbf{0.836 \pm 0.059}$ \\
\hline
\end{tabular}%
}
\end{table}

\begin{table}
\caption{Ablation study: AUROC of view assignment and augmentation variants}
\label{tab:ablation}
\renewcommand{\arraystretch}{1.5}
\setlength{\tabcolsep}{8pt}
\resizebox{\textwidth}{!}{%
\begin{tabular}{|l|c|c|c|c|c|}
\hline
\large\textbf{Trained on} &
\large\textbf{No Asymmetry} &
\large\textbf{Common Views} & \large\textbf{Reverse Views} &
\large\textbf{With Vessel} & \large\textbf{Our Approach} \\
\hline
\large REFUGE & \large $0.772 \pm 0.106$ & \large $0.799 \pm 0.075$ & \large $0.741 \pm 0.100$ 
& \large $0.812 \pm 0.068$ & \large $\mathbf{0.818 \pm 0.046}$ \\
\large AIROGS & \large $0.809 \pm 0.068$ & \large $0.805 \pm 0.062$ & \large $0.749 \pm 0.081$ 
& \large $0.831 \pm 0.063$ & \large $\mathbf{0.836 \pm 0.059}$ \\
\hline
\end{tabular}%
}
\end{table}

\subsection{Evaluation} To evaluate generalization, we train a linear classifier on frozen CLS token embeddings, which encode linearly separable, classification-relevant features in ViT-based models, and evaluate across \textit{seven glaucoma benchmarks} spanning diverse imaging conditions and populations~\cite{acrima,fives,papila,chaksu,lag,refuge,origa}. LAG, ORIGA, and FIVES overlap with the pretraining corpus, but since SSL uses no label information, this overlap introduces at most a transductive bias, meaning the model has implicitly encountered the unlabeled test image manifold during pretraining, similar to transductive learning, rather than any genuine task-specific supervision or label leakage. We use \textit{two linear-probe regimes}: a low-data setting using the REFUGE~\cite{refuge} training split ($\sim$1.2k images) and a large-scale setting using AIROGS~\cite{airogs} ($\sim$78k images, quality-filtered via QuickQual~\cite{engelmann2023quickqual}). We compare against two baseline categories: SSL approaches, comprising Vanilla DINO trained on our pretraining dataset and RETFound-DINOv2~\cite{retfound} pretrained on 1.6M unlabeled fundus images, and CLIP-based vision-language models, comprising RetiZero~\cite{retizero} and FLAIR~\cite{flair}, which learn joint image-text representations from expert-annotated pairs. Comparison against SSL baselines measures our augmentation strategy's effectiveness, while comparison against CLIP-based models, known for strong few-shot transferability, contextualizes our approach relative to retinal VLMs that had access to supervisory signal during pretraining.

\label{experiments}

Table~\ref{tab:results} reports AUROC with 95\% confidence intervals (CIs) via DeLong's method across seven datasets under both linear-probe regimes, with the best result per dataset in bold and the final row summarizing mean $\pm$ standard deviation. Compared against SSL baselines, our approach outperforms both Vanilla DINO and RETFound in mean AUROC across both regimes (0.818 vs. 0.788/0.768 and 0.836 vs. 0.780/0.801), confirming that task-aware augmentation yields stronger glaucoma representations than generic SSL pretraining. Among CLIP-based models, RetiZero performs competitively on most datasets but fails to generalize on CHAKSU, absent from its pretraining distribution, where our approach achieves notably higher scores (0.784 vs. 0.730 and 0.836 vs. 0.796). However, our approach trails RetiZero on REFUGE in the low-data regime, where large-scale supervised pretraining confers a stronger advantage. More broadly, our approach achieves the best worst-case AUROC across both regimes (0.757 and 0.787), avoiding the single-dataset failures observed in RETFound and RetiZero.

\subsection{Ablation Studies}
\label{ablation}
We validate four design choices in our approach through targeted ablations:

\begin{itemize}

\item \textbf{Reverse Views:} Our approach assigns task-relevant crops to the teacher and full-image crops to the student. This ablation reverses the assignment: the student receives task-relevant crops, the teacher full-image crops.

\item \textbf{Common Views:} In standard DINO, the teacher's crops are also passed to the student; we withhold them to avoid shortcut feature learning from highly correlated crops. This ablation additionally passes both teacher crops to the student, yielding five student crops in total.

 \item \textbf{No Asymmetry:} To validate our asymmetric design, we train a vanilla DINO variant with all local crops restricted to multi-scale optic disc crops with vessels, following the design principle similar to DINO-LG~\cite{dinolg}.

\item \textbf{With Vessel:} We ablate the effect of retaining vessels in the teacher crop, keeping all other design choices unchanged.

\end{itemize}

Table~\ref{tab:ablation} reports mean ± standard deviation AUROC across the same seven datasets for each ablation, with classifiers trained on REFUGE and AIROGS. \textbf{Reverse Views} causes the largest drop in both regimes (0.741, 0.749), confirming that the task-relevant crop must go to the teacher, not the student. \textbf{Common Views} also underperforms (0.799, 0.805), supporting our argument against giving the student the teacher's crop. \textbf{No Asymmetry} trails as well (0.772, 0.809), showing that restricting crops to task-relevant regions alone is insufficient without the teacher-student split. \textbf{With Vessel} performs closest to our full approach, with the largest gap on PAPILA, where vessel removal yields a significant improvement in the low-data regime ($\Delta$AUROC $= +0.045$, $p = 0.002$, DeLong's test) but not the large-scale regime ($\Delta$AUROC $= +0.029$, $p = 0.057$), indicating that the linear classifier is more sensitive to teacher-side nuisance variation under data-scarce conditions.

\section{Conclusion}
In this work, we analyzed the limitations of DINO's default multi-crop strategy for fundus images and developed an InfoMin-guided augmentation methodology that constructs task-aware optimal views, validated on color fundus photography for glaucoma classification. As the task-relevant crop is currently defined around the optic disc, extending this framework to retinal diseases with more distributed pathology and to additional modalities is a promising direction for future work.

\subsubsection{Acknowledgment.}
We gratefully acknowledge the funding support from \textbf{ZEISS India} and \textbf{Kotak-IISc AI/ML Centre}. We also acknowledge the use of large language models (LLMs) for improving the quality of presentation.

\subsubsection{Disclosure of Interests.} The authors have no competing interests to declare that
are relevant to the content of this article.


{\small
\bibliographystyle{splncs04}
\bibliography{references}
}
\end{document}